\documentclass[letterpaper]{article} 
\usepackage{aaai2026}  
\usepackage{times}  
\usepackage{helvet}  
\usepackage{courier}  
\usepackage[hyphens]{url}  
\usepackage{graphicx} 
\usepackage{natbib}  
\usepackage{caption} 
\usepackage{algorithm}
\usepackage{algorithmicx}
\usepackage{amssymb}
\usepackage{amsmath}
\usepackage{booktabs}
\usepackage{multirow}
\usepackage{pifont}

\usepackage{newfloat}
\usepackage{listings}
\DeclareCaptionStyle{ruled}{labelfont=normalfont,labelsep=colon,strut=off} 
\floatstyle{ruled}
\newfloat{listing}{tb}{lst}{}
\floatname{listing}{Listing}
\title{Rethinking Label Consistency of In-Context Learning:\\ An Implicit Transductive Label Propagation Perspective}
\author{
    Haoyang Chen\textsuperscript{\rm 1,\rm 2}, 
    Richong Zhang\textsuperscript{\rm 2,\rm 3 }\thanks{Corresponding Author}, 
    Junfan Chen\textsuperscript{\rm 1,\rm 2 }
}
\affiliations{
    \textsuperscript{\rm 1}School of Software, Beihang University, Beijing, China\\
    \textsuperscript{\rm 2}CCSE, School of Computer Science and Engineering, Beihang University, Beijing, China\\
    \textsuperscript{\rm 3}Zhongguancun Laboratory, Beijing, China

    Cauchy20011214@buaa.edu.cn,\{zhangrc, chenjf\}@act.buaa.edu.cn

}

\usepackage{bibentry}

\begin{document}

\maketitle

\begin{abstract}
Large language models (LLMs) perform in-context learning (ICL) with minimal supervised examples, which benefits various natural language processing (NLP) tasks. One of the critical research focus is the selection of prompt demonstrations. Current approaches typically employ retrieval models to select the top-K most semantically similar examples as demonstrations. However, we argue that existing methods are limited since the label consistency is not guaranteed during demonstration selection. Our cognition derives from the Bayesian view of ICL and our rethinking of ICL from the transductive label propagation perspective. We treat ICL as a transductive learning method and incorporate latent concepts from Bayesian view and deduce that similar demonstrations guide the concepts of query, with consistent labels serving as estimates. Based on this understanding, we establish a label propagation framework to link label consistency with propagation error bounds. To model label consistency, we propose a data synthesis method, leveraging both semantic and label information, and use TopK sampling with Synthetic Data (TopK-SD) to acquire demonstrations with consistent labels. TopK-SD outperforms original TopK sampling on multiple benchmarks. Our work provides a new perspective for understanding the working mechanisms within ICL.
\end{abstract}

\begin{links}
    \link{Code}{https://github.com/Cauchy2001/TopK_SD}
\end{links}

\section{Introduction}

Large language models (LLMs)~\cite{min2023recent,liu2023pre} demonstrate remarkable generative capabilities and achieve superior performance across a wide spectrum of traditional NLP tasks~\cite{mann2020language}, including sentiment analysis, text classification, and machine translation. 
The efficacy of these models is significantly enhanced by in-context learning (ICL)~\cite{dong2022survey,bai2024transformers}, which enables task execution through in-context inference with minimal supervised prompts. Compared to conventional fine-tuning approaches~\cite{devlin2018bert}, ICL accomplishes comparable task performance while substantially reducing data annotation costs~\cite{lester2021power}, primarily by eliminating the need for extensive parameter updates. For instance, by providing only 4-8 labeled examples as demonstrations, ICL allows models like GPT~\cite{radford2019language,floridi2020gpt,achiam2023gpt} and Llama~\cite{touvron2023llama,touvron2023llama2,dubey2024llama} to generalize to unseen tasks without explicit parameter updates and accomplishes comparable task performance while substantially reducing data annotation costs~\cite{lester2021power}.
For instance, by providing only 4-8 labeled examples as demonstrations, ICL allows models like GPT~\cite{radford2019language,floridi2020gpt,achiam2023gpt} and Llama~\cite{touvron2023llama,touvron2023llama2,dubey2024llama} to generalize to unseen tasks without explicit parameter updates and accomplishes comparable task performance while substantially reducing data annotation costs~\cite{lester2021power}.

\begin{table}[t]
	\centering
    
        \setlength{\tabcolsep}{2.5pt}
	\begin{tabular}{ccccc}
		\toprule
		Query & Demonstrations & Consistency &  Accuracy & Selection\\
		\midrule
		\multirow{2}*{\texttt{\color{green}{P}}} & \texttt{\color{green}{P}}\texttt{\color{green}{P}}\texttt{\color{green}{P}}\texttt{\color{green}{P}}\texttt{\color{green}{P}}\texttt{\color{green}{P}}\texttt{\color{red}{N}}\texttt{\color{red}{N}} & \color{green}{75\%} &  \color{green}{\(\uparrow\)} & \color{green}{\checkmark}\\
        &  \texttt{\color{green}{P}}\texttt{\color{green}{P}}\texttt{\color{red}{N}}\texttt{\color{red}{N}}\texttt{\color{red}{N}}\texttt{\color{red}{N}}\texttt{\color{red}{N}}\texttt{\color{red}{N}} &  \color{red}{25\%} & \color{red}{\(\downarrow\)}  & \color{red}{\ding{55}}\\
		\bottomrule
	\end{tabular}
	\caption{ On SST-2, for a query with the label "\texttt{\color{green}{positive}}", our method selects demonstrations with the \texttt{\color{green}{same}} label to improve ICL performance. }
\label{tab:dataset}
\end{table}

The performance of ICL critically hinges on demonstration selection ~\cite{sorensen2022information,zhang2022active,li2023finding}, their ordering~\cite{lu2022fantastically,liu2024lets}, and formatting~\cite{kim2022self,hao2022structured}; hence, selecting appropriate demonstrations for ICL in large models is paramount.
Consequently, the research community has introduced a wealth of methods for selecting demonstrations while also selecting demonstrations at the corpus or instance level with key factors such as diversity and similarity~\cite{luo2024context,dong2022survey}.
A common retrieval approach generates embeddings~\cite{reimers2019sentence}, calculates Euclidean distance or cosine similarity, and selects the top-K most similar demonstrations~\cite{liu2021makes}. 
The nearest-neighbor examples' labels enhance the generative model’s final prediction.
Subsequent work has extended TopK method, including MDL~\cite{wu2022self} and ConE~\cite{peng2024revisiting}. 
The effectiveness of these methods is dependent on TopK retrieval.

To understand the mechanisms of ICL, some studies have focused on influential factors, including data distribution~\cite{wies2024learnability}, training process~\cite{ding2023causallm}, and diversity~\cite{an2023context}, among others. Currently, there is no clear explanation for why ICL works. Existing principled explanations for ICL include the Bayesian veiw~\cite{xie2021explanation,wies2023learnability}, which suggests that demonstrations of ICL are used to activate the latent concepts learned during pretraining, and the gradient descent view~\cite{dai2022why,oswald2023transformers}, which posits that the effects of ICL are equivalent to those of fine-tuning. The work~\cite{wang2023label} interpret ICL's working principle via transformers' attention mechanism, arguing that label words anchor demonstration information to form final predictions.

Based on the existing research on the interpretability of ICL, most studies consider that ICL is not conventional learning~\cite{kossen2023context}. However, if we rethink from the perspective of learning paradigms, ICL should be considered transductive learning, rather than inductive learning. From Bayesian view, we introduce the latent concepts into the transductive optimization objective of ICL. Under the assumptions of Bayesian inference framework, we deduce the working principle of latent concepts in ICL. Latent concepts help map sentences to labels. Moreover, we derive that similar demonstrations can effectively guide the latent concepts corresponding to the query, and consistent labels can estimate whether the guided concepts related to the query. Therefore, we establish a transductive label propagation framework to explain how demonstrations in ICL propagate concepts to query and use label consistency as a lower bound for evaluating propagation error. The derivation reveals why label consistency is important.

Since we do not have sufficient prior knowledge of the query labels during demonstration selection, and existing methods do not incorporate labels into the selection, we designed a data synthesis method to synthesize‌ new embeddings with label information. We sample from the synthesized embeddings with TopK method called TopK with Synthetic Data (TopK-SD). This method can better select demonstrations with consistent labels. Compared with TopK with original embeddings, TopK-SD shows significant improvement on multiple benchmarks. The reason is that TopK-SD maintains semantic similarity  and greatly improves label consistency. Some methods based on TopK also conducted ablation studies, replacing the TopK module with the TopK-SD module, and it demonstrates the effectiveness of TopK-SD. It has verified the importance of label consistency and corroborated derivation of Bayesian transduction.

Briefly, the contribution of this study can be summarized  as follows: (1) We rethink ICL as a form of transductive learning, derive the roles of demonstrations and labels through Bayesian view, model transductive label propagation framework, and highlight the importance of label consistency. (2) We propose a data synthesis method using semantic and label information, apply TopK-SD sampling to get consistent-label demonstrations. Experiments show that TopK-SD outperforms TopK with higher accuracy.
\section{Related Works} 
\noindent{\bf In-Context Learning} As an emerging learning paradigm, In-Context Learning (ICL)~\cite{dong2022survey,bai2024transformers} has attracted significant attention in the field of natural language processing (NLP) in recent years~\cite{mann2020language}. ICL enables models to perform inference directly using a small number of labeled examples (i.e., demonstrations) provided in the context, without requiring large-scale parameter updates~\cite{devlin2018bert}. Compared to traditional fine-tuning methods, ICL achieves comparable task performance while substantially reducing the cost of data annotation~\cite{lester2021power}. It makes ICL particularly advantageous in low-resource scenarios.

\noindent{\bf Interpretability of ICL} Despite ICL's remarkable performance in practice, its underlying mechanisms remain unclear. Some studies have approached ICL from Bayesian view, suggesting that demonstrations activate latent concepts learned during pretraining~\cite{xie2021explanation,wies2023learnability}. Others have explained ICL through the view of gradient descent, positing that its effects are equivalent to fine-tuning~\cite{dai2022why,oswald2023transformers}. The work~\cite{wang2023label} have explored the attention mechanisms of transformers, arguing that labels aggregate demonstration information for label prediction. While these studies offer different perspectives on understanding mechanisms, many questions still remain open for investigation.

\noindent{\bf Demonstration Selection of ICL} ICL critically hinges on the demonstration selection ~\cite{sorensen2022information}, their ordering~\cite{lu2022fantastically}, and formatting~\cite{kim2022self}. Base above factors, demonstration selection methods can generally be divided into corpus-level and instance-level approaches. Corpus-level methods include Votek~\cite{su2022selective},Q-Learning~\cite{zhang2022active}, and others. However, corpus-level methods often perform worse than instance-level methods, which is why most researchers focus more on the latter. Instance-level methods include TopK, which select the top-K most similar demonstrations by calculating cosine similarity~\cite{liu2021makes}. Subsequent work has extended the TopK method, such as MDL~\cite{wu2022self} and ConE~\cite{peng2024revisiting}. These methods significantly improve ICL's performance by optimizing the retrieval process and demonstration selection strategies. However, their effectiveness heavily relies on TopK retrieval.
\section{Problem Formulation}
In-context learning is a training-free paradigm that enables LMs to learn downstream tasks using only a few demonstrations. Therefore, ICL can also be regarded as few-shot learning in such scenarios~\cite{mann2020language}. Formally, for an input query \( x \), there is a label candidate set \( \mathcal{Y}=\{y_1, \cdots, y_M\} \). 
The likelihood of a candidate label y is derived from a scoring function \( f \) computed by LM within the context \( C = \{x_1, y_1, \dots, x_k, y_k\} \) ~\cite{dong2022survey}. 
\begin{equation}
    P\left(y \mid x\right) \triangleq f_{LM}\left(y \mid C, x\right) = f_{LM}\left( y \mid X,Y,x\right) 
    \label{eq:define}
\end{equation}
Context \( C \) is a prompt constructed via \( k \)-shot learning, where each pair \( (x_i, y_i) \) represents the sentence and label of the \( i \)-th selected example. The sentences of demonstrations in context is \( X = \{x_i\}_{i=1}^k \) and the labels \( Y = \{y_i\}_{i=1}^k \). Obviously, \( C = \{X, Y\} \). The final predicted label $\hat{y}$ is the label with highest probability.
\begin{equation}
    \hat{y} = \arg\max_{y \in \mathcal{Y}} \, P\left(y \mid x\right)
    \label{eq:prediction}
\end{equation}
The task is to select demonstrations \(\{(x_i,y_i)\}_{i=1}^k\) that enhances ICL and improves the prediction performance.
\section{Methodology}
In this section, we rethink ICL from the perspective of transductive learning and argue that it can be modeled as label propagation. We explain the role of labels within ICL. Building on these insights, we introduce a data-synthesis method that leverages both demonstration semantics and label information to synthesis new embeddings and employ TopK for sampling on them. This method is referred to as TopK with Synthesis Data (TopK-SD).
\subsection{Rethink Learning Paradigms of ICL}
ICL, as a capability of LMs, can learn from only a few examples. While the learning mechanism behind it is still not fully understood by researchers, many views have been proposed to view ICL, such as the Bayesian view ~\cite{xie2021explanation,wies2023learnability} and gradient descent view~\cite{dai2022why,oswald2023transformers}. In existing research work, it is considered that ICL is not conventional learning~\cite{kossen2023context}. From the perspective of learning paradigms, we update the view that ICL is regarded as transductive learning, rather than inductive learning in traditional sense. 

\noindent{\bf Transductive Learning} Transductive learning uses the training set \( \mathcal{D}_{\text{train}} = \{(x_i, y_i)\}_{i=1}^{n} \) and the unlabeled test set \( \mathcal{X}_{\text{test}} = \{x_j\}_{j=1}^{m} \) to directly infer test labels without learning a generalizable function. It focuses on local optimization rather than generalization, unlike inductive learning which aims to predict unseen data. Mathematically:
\begin{equation}
    \hat{y}_{1:m} = \arg\!\!\!\!\max_{y_{1:m}\in\mathcal{Y}^m} \!\!\!\! P \left(y_{1:m} \mid \mathcal{D}_{\text{train}}, \mathcal{X}_{\text{test}}\right)
    \label{equ:trans}
\end{equation}

Revisiting Equation \ref{eq:define}, it's significant that the goal of demonstration selection should be to optimize the function \(f_{LM}\). And while \( m = 1 \), the form of Equation \ref{equ:trans} is identical to that of the function \( f_{LM} \). Accordingly, we provide the optimization of ICL with the definition of transductive learning.
\begin{equation}
    \begin{aligned}
    \hat{y} = \arg\max_{y \in \mathcal{Y}}  P \left( y \mid X, Y, x \right)
    \end{aligned}
    \label{equ:trans}
\end{equation}
Reconsidering the nature of ICL, it should be fundamentally regarded as a transductive learning paradigm.

\begin{figure*}[ht]
	\begin{center}
		\begin{tabular}{c}
                \hspace{-.27cm}
		    \includegraphics[width=\textwidth]{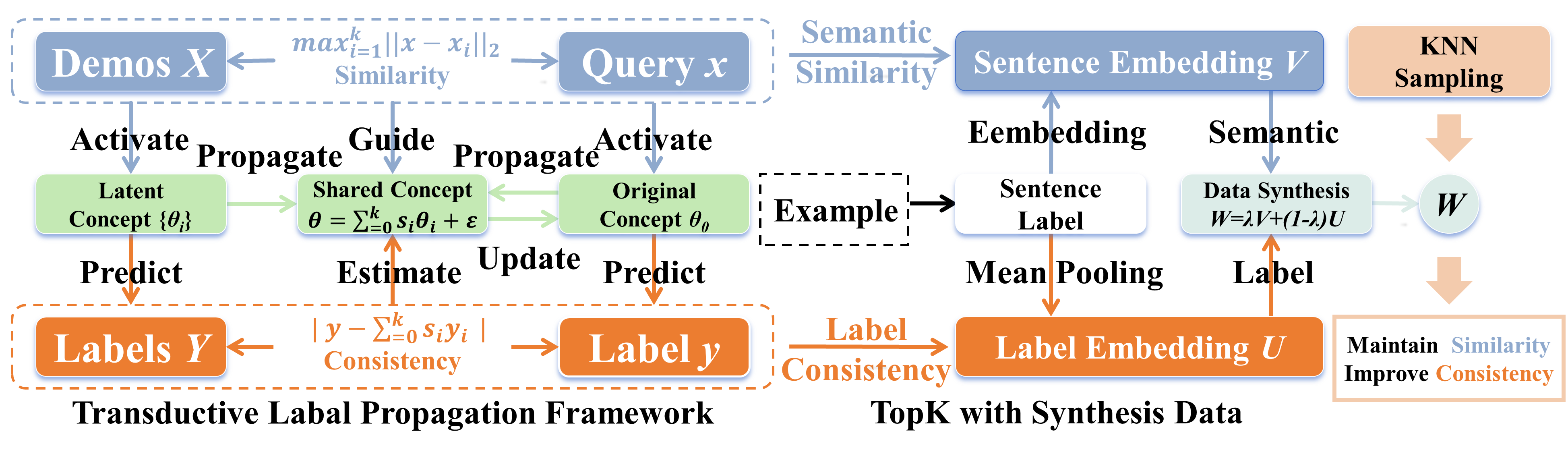}
		\end{tabular}
	\end{center}
	\caption{TopK sampling is performed on the synthesized embeddings with semantic and label information.}
	\label{fig:structure}
\end{figure*}
\subsection{Bayesian Transduction} 
We derive the transductive learning objective of ICL. Express the transductive learning optimization Function \ref{equ:trans} in terms of conditional probabilities as Equation \ref{equ:condition}. It shows that \( P\left(y \mid X, Y, x\right) \) is proportional to \( P\left(Y, y \mid X, x\right) \).
\begin{equation}
     P \left( y \mid X, Y, x \right) = \frac{P \left(  Y , y \mid  X, x\right)}{P \left(   Y \mid  X, x\right)} \propto P \left(  Y , y \mid  X, x\right)
    \label{equ:condition}
\end{equation}
Combining the Bayesian inference framework hypothesis, make further deduction for ICL transductive learning.

\noindent{\bf Bayesian Inference Framework} In the Bayesian framework, ICL is explained as implicit Bayesian inference shown as Equation \ref{eq:bayes}~\cite{xie2021explanation}.
\begin{flalign}
\begin{split}
P \left(y \mid X,Y, x\right) = \int_{\theta} p(y \mid \theta, X,Y, x) \, p(\theta \mid X,Y, x) \, \mathrm{d} \theta \\
= \int_{\theta_0,\theta_1, \ldots, \theta_k} p(y \mid x_1, y_1, \theta_1, \ldots, x_k, y_k, \theta_k, x, \theta_0) \\
\quad \cdot p(\theta_0 ,\theta_1,, \ldots, \theta_k  \mid X, Y, x)  \, \mathrm{d}\theta_0 \, \mathrm{d}\theta_1  \, \cdots \, \mathrm{d}\theta_k
\end{split}
\label{eq:bayes}
\end{flalign}
The role of demonstrations is to activate the latent concept \(\theta\) which characterizes the relationship between sentences and labels and helps LM to better understand the task intentions. Further research indicates that there is a latent concept \(\theta_i\) related to $i^{th}$ demonstration, \(\theta_0\) is raleted to query \( x \). Moreover, \(\theta\) represents the shared concept of all \(\theta_i\) and \(\theta_0\) and characterizes the overall mapping relationship in ICL. The mechanism of ICL is to activate the knowledge concepts learned during the pre-training stage based on the demonstrations in the context, and to generate answers \(y\) for the problem \(x\) according to these concepts.

Incorporating the Bayesian inference framework, latent concept \( \theta \) is introduced to determine the mapping \(f_{\theta}: X {\rightarrow} Y \). The mapping \( f_{\theta} \) is also applicable from \( x \) to \( y \), which is referred to as ICL. So \(P\left(Y,y \mid X,x\right)\) characterizes such a mapping \( f_{\theta} \). Applying the Bayesian inference framework to \(P\left(Y,y \mid X,x\right)\), Equation \ref{equ:theta} can be described by introducing the latent concept \( \theta \) as follows:
\begin{align}
P \left( Y, y \mid X, x \right) =    \int_{\theta} p\left( Y, y \mid \theta, X, x \right)  p\left(\theta \mid X, x\right) \, \mathrm{d}\theta 
\label{equ:theta}
\end{align}

\noindent{\bf Latent Optimization Objective} Substituting Equation \ref{equ:theta} into Equation \ref{equ:condition} yields Equation \ref{equ:condition2}. Since the denominator is a normalization constant, while the numerator represents the exploration of the latent concept \(\theta\), the fundamental objective of ICL should be to find suitable \(\theta\) related to query \(x\) and its label \(y\).
\begin{align}
P \left( y \mid X, Y, x \right) =  \frac{\int_{\theta} p \left( Y, y \mid \theta, X, x \right)  p\left(\theta \mid X, x\right) \, \mathrm{d}\theta}{\int_{\theta} \int_{y} p \left( Y, y \mid \theta, X, x \right)  p\left(\theta \mid X, x\right) \, \mathrm{d}y\mathrm{d}\theta}
    \label{equ:condition2}
\end{align}

\subsection{Label Propagation Framework}
Based on Bayesian transduction, the learning mechanism of ICis elaborated and a label propagation framework is established to explain ICL.

\noindent{\bf Bayesian Belief for Concept \(\theta\)} From Bayesian view, the selection of demonstrations \(X\) is an update to the concept \(\theta\). Equation \ref{equ:meaning} express \( p\left(\theta \mid X, x\right) \) in the form of Bayes' theorem. Here, it reveals that LM has a prior probability \( p\left(\theta \mid x\right) \) regarding \(\theta\) while $x$ is known. As \( X \) is introduced as evidence for the likelihood estimation of \(\theta\), LM updates the posterior probability \( p\left(\theta \mid X, x\right) \). Since query \( x \) is known, and demonstrations \( X \) need to be selected, if \( X \) is highly similar to \( x \), it can help the model better estimate the shared concept \(\theta\) related to \( x \). This explains from Bayesian belief why similarity is important for ICL~\cite{liu2021makes}. 
\begin{align}
  p\left(\theta \mid X, x\right) =\frac{p\left(X \mid \theta,x\right)p\left(\theta \mid x\right)}{p\left(X\mid x\right)} 
  \label{equ:meaning}
\end{align}

In addition, \( p\left(Y, y \mid \theta ,X, x\right) \) in Equation \ref{equ:theta} can be regarded as the likelihood estimate of the shared concept \( \theta \) given demonstrations' labels \( Y \) and query's label \( y \), while \( X \) and \( x \) are known. When \( Y \) and \( y \) are as consistent as possible, \( p\left(Y, y \mid \theta ,X,x\right) \) can offer a better estimate of \( \theta \). If the consistency between \( Y \) and \( y \) is higher, it can better reflect that the shared concept \(\theta\) is more strongly associated with \( x \).
From this view, the goal of ICL should be intrinsically linked to the concept associated with the query,and we introduce label propagation to explain the update of concepts.

\noindent{\bf Transductive Label Propagation } Transductive label propagation is a semi-supervised learning approach that uses the labels of labeled data to infer the labels of unlabeled data. Based on Equation \ref{equ:theta}, the similarity of demonstrations for guiding latent concepts and the consistency of labels for estimating latent concepts align with the smoothness assumption of label propagation: similar data tend to have the same label. Base on this, we propose the view that ICL is a label propagation framework to update the concept \(\theta\).

\noindent{\bf Update of Concept \(\theta\)} The feature propagation in label propagation, as shown in Equation \(\ref{equ:feature2}\), represents the update of original concept \(\theta_0\) of \(x\) with the incorporation of \(\{\theta_i\}_{i=1}^k\), resulting in shared concept \(\theta\).
\(s_i\) represents propagation coefficient, which is related to factors such as the semantic similarity between \(x\) and \(x_i\), the position of the \(x_i\) in order of the context and other relevant factors. Therefore, Equation \ref{equ:feature2} can be regarded as further updating information of input query by incorporating in-context demonstrations.

There are apparently errors, so it is necessary to introduce an error term into the feature propagation Equation \ref{equ:feature2}. \( \epsilon \) denotes the propagation error within the LM in ICL. 
\begin{equation}
    \mathbf{\theta} = \sum_{i=0}^{k} s_i \mathbf{\theta}_i +\boldsymbol{\epsilon}
    \label{equ:feature2}
\end{equation}

The update of \( \theta \) also signifies the update of the label \( y \), in an effort to characterize the propagation error using \( y \).

\noindent{\bf L-Lipschitz Constraint} The mapping of ICL \(f_{\theta}: X {\rightarrow} Y \) is smooth and satisfies the L-Lipschitz constraint as follow:
\begin{equation}
    \left| \mathbf{y}_1 - \mathbf{y}_2 \right| = \left| f_{\theta}\left(\mathbf{x}_1\right) - f_{\theta}\left(\mathbf{x}_2\right) \right| \leq L \left\| \mathbf{x}_1 - \mathbf{x}_2 \right\|_2
\end{equation}
The propagation error can be characterized by the constraint.
\begin{equation}
    \left| \mathbf{y} - \sum_{i=0}^{k} s_i \mathbf{y}_i \right| \leq L \|\boldsymbol{\epsilon}\|_2 + o\left( \max_{i=1}^{k} \| \mathbf{x} - \mathbf{x}_i \|_2 \right)
    \label{equ:lower}
\end{equation}
\noindent{\bf Estimation of Propagation Error} The L-Lipschitz constraint provides a lower bound for the error estimate. \( L\|\boldsymbol{\epsilon}\|_2 \) represents the noise caused by propagation error \(\epsilon\). Noise \( L\|\boldsymbol{\epsilon}\|_2 \) is proportional to the magnitude of error \(\epsilon\), and physical significance can be understood as an amplification of error \(\epsilon\). \(o (\max_{i=1}^{k} \| \mathbf{x} - \mathbf{x}_i \|_2 )\) represents the asymptotic upper bound of the maximum difference between \(x\) and \(x_i\), reflecting semantic similarity. In addition, \(| \mathbf{y} - \sum_{i=0}^{k} s_i \mathbf{y}_i |\) describes the discrepancy  between the label obtained through propagation and the ground-truth label. It is worth noting that \( y_0 \) describes the initial label assigned to \( x \) by LM.

Noise's lower bound is determined by both semantic similarity and label consistency. Combining Equation \ref{equ:theta} and Equation \ref{equ:lower}, the similarity demonstrations are intended to reduce the propagation error, while label consistency is aimed at estimating the error. This provides the two guiding principles for demonstration selection. The process should take into account both of these factors. In other words, base on the label propagation framework, ICL should focus more on demonstrations, which should have a high semantic similarity to the query and share the same label.

\subsection{Label Consistency of Data Synthesis}

In ICL, no substantial prior knowledge is available for the labels of input queries and most methods do not consider the utilization of label information of demonstrations. To address this, we design a data synthesis method based on semantic similarity measured by embedding models, which facilitates sampling demonstrations with high semantic similarity and the same label from the synthesized embeddings.

\noindent{\bf Embedding Interpolation} To obtain demonstration with consistent label, we employ an interpolation method to force vectors of the same category to converge towards a central vector. Let \(\mathcal{D}=\{(x_i, y_i)\}_{i=1}^{n}\) denote candidate demonstration set and \(\mathcal{Y}\) as the label set. \(\mathbf{V}_i\) denote the embedding of the \(i^{\rm th}\) demonstration's sentence. \(\mathbf{U}_k\) denote the center vector for category $k$, which is defined as the mean-pooling of all embedding vectors belonging to category $k$ as follow:
\begin{equation}
\label{eq:center}
\mathbf{U}_k = \frac{
\sum_{i=1}^{n} \mathbf{V}_{i} \cdot \mathbb{I}\left[y_i = k\right]
}{
\sum_{i=1}^{n} \mathbb{I}\left[y_i = k\right]
}
\end{equation}

\(\mathbf{U}_k\) is the semantic center of category \(k\), that is, the label embedding representing the characteristics of the category. $\mathbb{I}[A]$ is an indicator function such that when Boolean expression $A$ is true, $\mathbb{I} = 1$; otherwise, $\mathbb{I} = 0$.
We interpolate to synthesize embedding using sentence and label embedding.
\begin{equation}
  \mathbf{W}_i = \lambda \mathbf{V}_{i} +\left(1-\lambda\right)\mathbf{U}_{y_i}
  \label{eq:embedding interpolation}
\end{equation}

The synthesized embeddings can better conform to the smoothness theory. In this way, our method can utilize label information to enhance the original data features.

For each input query, interpolation is performed following equation \ref{eq:embedding interpolation}. Since query lacks label information, we select an appropriate estimation equation to minimize the error between the estimated value and ground-truth value.
Define \(\mathbf{U}\) as the reference vector, which be obtained through the mean-pooling of all \(\mathbf{U}_k\).
\begin{equation}
\label{center}
\mathbf{U} = \frac{
\sum_{k \in \mathcal{Y}} \mathbf{U}_k
}{
\left |\mathcal{Y} \right|
}
\end{equation}
Selecting reference vectors without any semantic information can reduce bias towards any specific category. We perform the following interpolation transformation based on $\mathbf{U}$.
\begin{equation}
  \label{eq:estimate}
  \mathbf{W}_i = \lambda \mathbf{V}_{i} +\left(1-\lambda\right)\mathbf{U}
\end{equation}

By employing such data synthesis method, the synthetic embeddings of demonstrations from the same category can achieve better clustering performance, and the nature of the synthetic data better conforms to the smoothness theory.

\noindent{\bf Demonstration Selection} TopK method samples K-Nearest Neighbors (KNN) on the original embeddings to serve as demonstrations for LM to perform ICL. The advantage of this approach is that it allows for the selection of the K most semantically similar samples as demonstrations, but the label consistency between demonstrations and the query is not very good. We propose a new method, TopK with data synthesis (TopK-SD), which selects KNN from the synthesized embeddings to serve as demonstrations for ICL. This method not only maintains semantic similarity but also improves label consistency. The resulting labels from sampling exhibit significantly improved consistency. The propagation error of ICL can be better reflected through the labels of the selected demonstrations.

\section{Experiment} 

\begin{table*}[ht]
\centering
    \resizebox{\textwidth}{!}{%
\begin{tabular}{ccccccccccccc}
\toprule
 \textbf{Model} & \textbf{Method}  &\textbf{SST-2} & \textbf{SST-5}  &   \textbf{AGNews} &   \textbf{TREC} &  \textbf{CR} & \textbf{Subj} & \textbf{MNLI} & \textbf{QNLI} & \textbf{RTE} &  \textbf{Avg.}  &  \(\Delta\) \\
\midrule
\multirow{6}*{LlaMA3}  & Prompt & 86.3 & 26.8 & 70.5 & 47.0 & 83.8 & 61.4 & 47.7 & 50.0 & 65.0 & 59.8 & +21.3 \\
& Random & 95.7 & 47.4 & 82.4 & 66.4 & 89.4 & 90.9 & 56.8 & 58.4 & 72.9 & 73.4 & +7.8 \\
& Votek & 96.5 & 50.3 & 85.8 & 64.2 & 83.2 & 94.3 & 63.0 & 53.1 & 70.4 & 73.4 & +7.7\\
& BM25 & 95.5 & 47.4 & 93.0 & \bf{93.2} & 91.8 & 95.7 & 62.6 & 61.8 & \bf{74.4} & 79.5 & +1.6\\
& TopK & 96.0 & 52.7 & 93.5 & 91.0 & 91.5 & \bf{96.7} & 63.4 & 62.7 & 70.4 & 79.8 & +1.4\\
& TopK-SD & \bf{96.5} & \bf{53.7} & \bf{93.9} & 92.6 & \bf{92.8} & 96.3 & \bf{64.7} & \bf{67.0} & 72.6 & \bf{81.1} & - \\
\midrule
\multirow{3}*{GPT-j}  
& Random & 91.2 & 41.6 & 70.8 & 49.6 & 81.4 & 72.0 & 39.7 & 51.4 & 55.2 & 61.4 & +12.6 \\
& TopK & 93.8 & 49.2 & 90.1 & 88.6 & 89.6 & 92.0 & 43.0 & 52.8 & 53.8 & 72.5 & +1.5\\
& TopK-SD & \bf{94.6} & \bf{50.1} & \bf{90.6} & \bf{91.6} & \bf{90.7} & \bf{92.3} & \bf{44.2} & \bf{55.3} & \bf{57.0} & \bf{74.0} & - \\
\midrule
\multirow{3}*{LlaMA2}  
& Random & 92.1 & 46.1 & 82.6 & 63.2 & 90.7 & 47.1 & 50.3 & 56.5 & 69.7 & 66.5 & +10.6 \\
& TopK & 94.9 & 53.6 & 91.3 & 89.0 & 93.4 & 82.5 & \bf{52.2} & 59.5 & \bf{67.1} & 75.9 & +1.2\\
& TopK-SD & \bf{95.3} & \bf{53.7} & \bf{92.1} & \bf{92.2} & \bf{93.6} & \bf{85.9} & 51.3 & \bf{63.1} & 66.8 & \bf{77.1} & - \\
\midrule
\multirow{3}*{DeepSeek}  
& Random & 95.5 & 44.4 & 80.9 & 61.2 & 93.1 & 74.9 & 45.4 & 54.2 & 59.9 & 67.7 & +9.2 \\
& TopK & 95.8 & 51.8 & 92.1 & 87.8 & 93.4 & 93.5 & \bf{50.6} & 58.4 & \bf{66.1} & 76.6 & +0.3\\
& TopK-SD & \bf{95.8} & \bf{52.6} & \bf{92.1} & \bf{90.4} & \bf{93.9} & \bf{94.1} & 48.4 & \bf{61.1} & 63.5 & \bf{76.9} & - \\
\bottomrule
\end{tabular}
}
\caption{TopK-SD is compared with various baseline methods across four models and nine datasets. Best results are \textbf{bold}.}
\label{tab:main_results}

\end{table*}

\subsection{Datasets and Experiment Setting}

\noindent{\bf Dataset} We use widely-used datasets, comprising six classification tasks and three natural language inference (NLI) tasks~\cite{sun2023text}. The data includes SST-2~\cite{socher2013recursive}, SST-5\cite{socher2013recursive}, AGNews~\cite{zhang2015character}, TREC~\cite{voorhees2000building}, CR~\cite{hu2004mining}, Subj~\cite{pang2004sentimental}, MNLI~\cite{williams2017broad}, QNLI~\cite{rajpurkar2016squad} and RTE~\cite{rajpurkar2016squad}.

\noindent{\bf Experiment Setting}  We employ a sentence-transformer models as the retrieval model all-roberta-large-v1~\cite{reimers2019sentence}, along with the inference models GPT-j-6b~\cite{wang2021gpt}, LlaMA2-7b~\cite{touvron2023llama}, deepseek-llm-7b-base~\cite{bi2024deepseek} and LlaMA3-8b ~\cite{grattafiori2024llama}. We performed 8-shot experiments using A800 with this training set for demonstration selection, randomly sampling 1,000 test instances thrice to compute the average accuracy for three times.

\subsection{\ Baseline Methods}

We use TopK-SD method to generate embedding interpolations and sample KNN to select demonstrations and compare it with directly using prompt templates, evaluating their impact on ICL. 

\noindent{\bf Prompting } Prompting without in-context demonstrations. 

\noindent{\bf Random } Selecting demonstrations randomly. 

\noindent{\bf Votek~\cite{su2022selective} } Selection methods that are enlightening for ICL at the corpus level with voting method.

\noindent{\bf BM25~\cite{robertson2009probabilistic} } Using term frequency and inverse document frequency to evaluate document-query relevance and document length. 

\noindent{\bf TopK~\cite{liu2021makes,gao2020making} } Choosing the top-K most semantically similar demonstrations based on the embedding similarity.

In addition, for some methods based on the TopK approach, we replace the TopK module of the method with our TopK-SD method and verify the effectiveness of TopK-SD.

\noindent{\bf MDL~\cite{wu2022self} } Adopting a framework that first ranks combinations of demonstrations based on the Minimum Description Length (MDL) principle, and chooses the combination with the best MDL scores.

\noindent{\bf ConE~\cite{peng2024revisiting} } Using a framework that ranks demonstrations by Conditional Entropy (ConE), and chooses the top k with the best ConE scores.

\noindent{\bf DPP~\cite{ye2023compositional} } A method using Determinantal Point Processes (DPP) to model demonstration-input interactions and optimize via contrastive learning for selection.

\subsection{Text Classification Results}

\noindent{\bf Main Results }
In the experiments on LLaMA3, TopK-SD tested ten different values of \( \lambda \) (i.e., \( \lambda = 0.0, 0.1, \ldots, 0.9 \)), and the value that achieved the highest accuracy is selected as the evaluation result. In the experiments on GPT-J, LLaMA2, and DeepSeek, the parameter \( \lambda \) for TopK-SD is set to 0.7. The results of the TopK-SD method compared to the baseline method are shown in Table \ref{tab:main_results}. It demonstrates that in the experiments on the LLaMA3 model, selecting an appropriate \( \lambda \) for data synthesis enables TopK-SD to achieve the best performance on the validation set across different benchmarks. 
When \(\lambda\) is set to 0.9 on RTE, 0.8 on SST-2 and MNLI, 0.7 on SST-5 and Subj, 0.6 on TREC, 0.5 on AGNews, and 0.3 on QNLI.
Compared with TopK, the average accuracy of TopK-SD has increased by 1.4\%. Even on the other three models, where \( \lambda \) is set to only 0.7, most benchmarks and the average accuracy are improved. On three different models, the average accuracy of TopK-SD was improved by 1.5\%, 1.2\%, and 0.3\% compared to TopK, respectively. If appropriate values of \(\lambda\) are selected for the datasets, the accuracy of TopK-SD can be further improved.

\begin{table*}[ht]
\centering

\begin{tabular}{ccccccccccccc}
\toprule
 \textbf{Stage1} & \textbf{Stage2}  &\textbf{SST-2} & \textbf{SST-5}  &   \textbf{AGNews} &   \textbf{TREC} &  \textbf{CR} & \textbf{Subj} & \textbf{MNLI} & \textbf{QNLI} & \textbf{RTE} &  \textbf{Avg.}  &  \(\Delta\) \\
\midrule
 TopK & \multirow{2}*{MDL} & 96.7 & 54.1 & 87.4 & 87.0 & 93.6 & 93.6 & \bf{59.4} & 63.7 & 62.5 & 77.6 & +0.9 \\
TopK-SD &  & \bf{97.2} & \bf{55.9} & \bf{87.8} & \bf{90.2} & \bf{93.9} & \bf{94.6} & 58.8 & \bf{65.4} & \bf{62.5} & \bf{78.5} & - \\
\midrule
 TopK & \multirow{2}*{ConE} & 95.8 & 46.0 & \bf{92.2} & 93.6 & 91.5 & \bf{96.8} & \bf{58.9} & \bf{66.4} & 67.1 & 78.7 & +0.4 \\
TopK-SD &  & \bf{96.1} & \bf{46.1} & 91.9 & \bf{94.2} & \bf{92.6} & 96.5 & 58.7 & 65.8 & \bf{70.4} & \bf{79.1} & - \\
\midrule
 TopK & \multirow{2}*{DPP} & 95.4 & 49.9 & \bf{92.0} & 86.0 & 91.0 & 94.2 & 53.2 & 57.6 & \bf{69.7} & 76.5 & +0.6 \\
TopK-SD &  & \bf{96.1} & \bf{50.9} & 91.7 & \bf{86.8} & \bf{92.3} & \bf{94.3} & \bf{54.8} & \bf{59.5} & 67.5 & \bf{77.1} & - \\
\bottomrule
\end{tabular}
\caption{Replace the TopK module in Stage 1 of three methods with TopK-SD and conduct comparisons. Best results are \textbf{bold}.}
\label{tab:ablation_study}
\end{table*}

\begin{figure}[t]
\hspace{-.10cm}\includegraphics[width=\columnwidth]{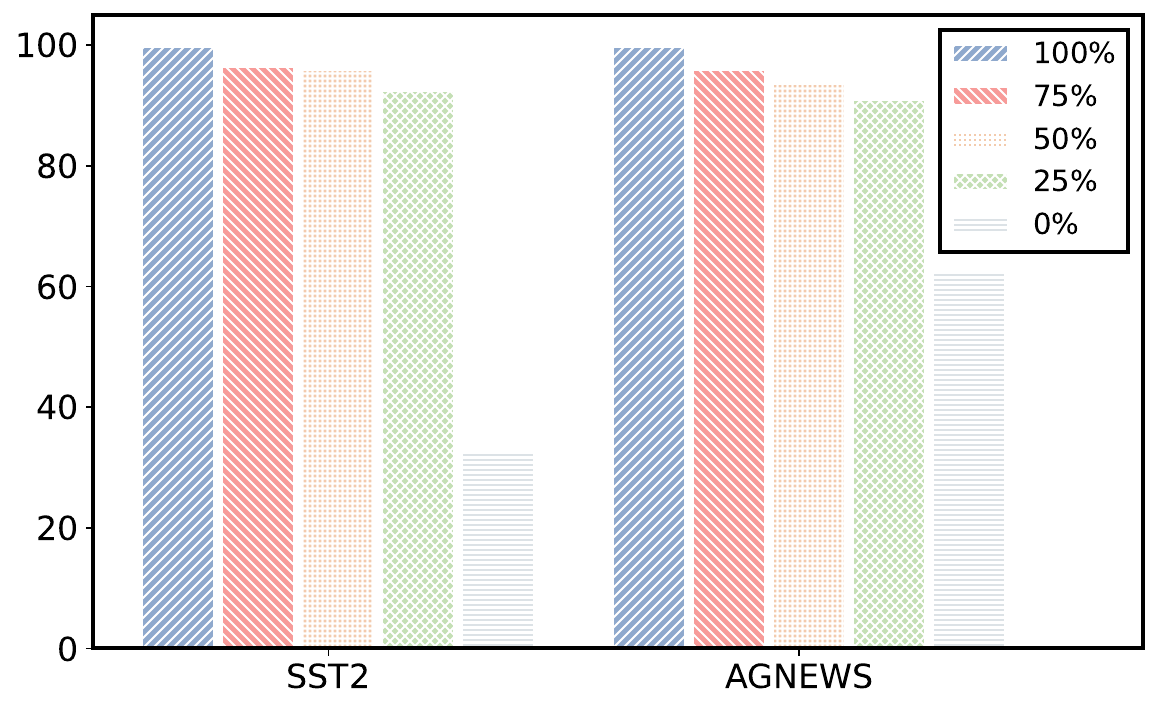}
  \caption{Investigate the relationship between label consistency and ICL accuracy on different datasets.}
  \label{fig:accuracy_consistency}
\end{figure}
\begin{figure}[t]
\hspace{-.10cm}\includegraphics[width=\columnwidth]{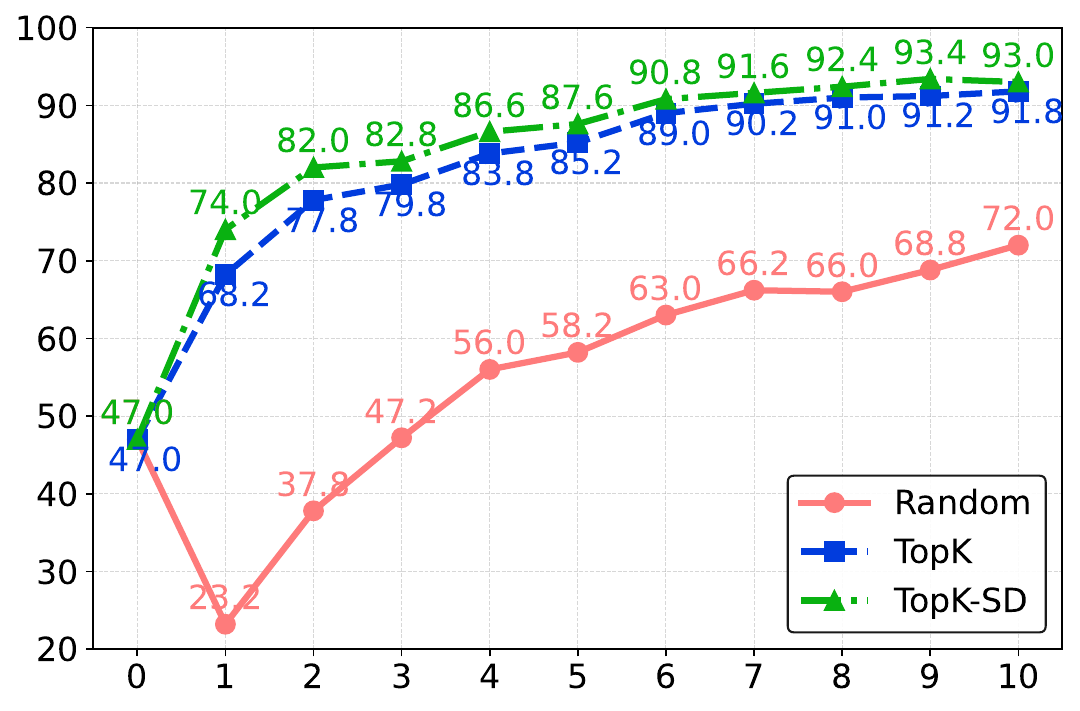}
  \caption{Using different numbers of demonstrations, compare TopK-SD (\(\lambda=0.7\)) and other methods on TREC.}
  \label{fig:numb}
\end{figure}

\noindent{\bf Ablation Study Results} In MDL, ConE, and DPP methods, retrieval involves two stages: Stage 1 uses TopK or TopK-SD (\(\lambda=0.7\)) to narrow down to 30 candidates, and Stage 2 employs three strategies to select 8 final demonstrations. We replaced TopK with TopK-SD and tested on LLaMA3. Results in Table \ref{tab:ablation_study} show that TopK-SD outperformed TopK with average accuracy gains of 0.9\%, 0.6\%, and 0.4\% across strategies, though the improvement is not significant, possibly due to the effective strategy in Stage 2 and the only choice of \(\lambda = 0.7\). Even so, it demonstrates that the narrowed candidate set by TopK-SD enhances label consistency, thereby improving performance across various scenarios.

\subsection{Analysis of Label Consistency in ICL}

\noindent{\bf Label Consistency} Since the propagation coefficient \(s_i\) is hard to characterize, label consistency is therefore quantified by the ratio of demonstrations with the same label. As depicted in Figure \ref{fig:accuracy_consistency}, the experimental outcomes demonstrate a positive correlation between label consistency and accuracy: as label consistency increases, the accuracy of ICL also improves. Furthermore, when no demonstration labels are consistent with \( y \), it implies a lack of guidance on how to extract the latent concept \( \theta \) that maps \( x \) to \( y \). Consequently, the accuracy significantly decreases. It is evident that label consistency contributes to ICL.

\noindent{\bf Demonstrations with Consistent Label} To ensure that the improvement in ICL due to enhanced label consistency is not from biased distribution of demonstration labels but from sufficient demonstrations providing necessary information for inference, we conducted ablation studies by removing inference module. For different \( \lambda \) values, KNN algorithm samples demonstrations on synthetic data. The vote method selects query labels by label-voting (without inference), while LM inferences for ICL based on the same demonstrations. Results in Table \ref{tab:vote} show that ICL significantly outperforms the vote method. It indicates that more demonstrations with the same label enable ICL to better uncover query-related latent concepts for inference rather than just copying labels. Additionally, we also compared the performance of three methods with different numbers of demonstrations, as shown in Figure \ref{fig:numb}. With few demonstrations, random method in few-shot learning performs worse than zero-shot learning. It often samples demonstrations with the same label difficultly, making it hard for ICL to classify accurately. Meanwhile, TopK-SD outperforms TopK when demonstrations are three or fewer, showing that label consistency is crucial with limited demonstrations.

\begin{table}[t]
\begin{tabular}{cccccc}\toprule
 \textbf{\(\lambda\)} & \textbf{Method}  &\textbf{SST-2} & \textbf{SST-5}  &   \textbf{AGNews} &   \textbf{TREC}  \\
 \midrule
 \multirow{2}*{0.0} & Vote & 49.5 & 17.6 & 25.0 & 18.8 \\
 & ICL & \bf{68.4} & \bf{18.4} & \bf{25.4} & \bf{18.8} \\
 
\midrule
  \multirow{2}*{0.2} & Vote & 90.6 & 17.7 & 42.4 & \bf{37.6} \\
& ICL & \bf{92.6} & \bf{18.6} & \bf{47.8} &  37.2 \\
\midrule
  \multirow{2}*{0.4} & Vote & 91.5 & 33.4 & 82.7 & 75.0 \\
& ICL & \bf{95.6} & \bf{43.3} & \bf{90.4} &  \bf{89.2} \\
\midrule
\multirow{2}*{0.6} & Vote & 90.6 & 43.3 & 91.3 & 84.6 \\
& ICL & \bf{96.1} & \bf{51.9} & \bf{93.3} &  \bf{92.6} \\
\midrule
\multirow{2}*{0.8} & Vote & 89.0 & 45.1 & 91.8 & 80.8 \\
& ICL & \bf{96.5} & \bf{53.3} & \bf{93.3} &  \bf{92.6} \\
\midrule
\multirow{2}*{1.0} & Vote & 87.3 & 42.6 & 91.7 & 76.8 \\
& ICL & \bf{96.0} & \bf{52.7} & \bf{93.5} &  \bf{91.0} \\
\bottomrule
\end{tabular}
\caption{In ablation study, inference process of LM is eliminated. With the same demonstrations collected, ICL is compared to label voting method. Best results are \textbf{bold}.}
\label{tab:vote}
\end{table}

\begin{figure}
\begin{tabular}{cc}
      \hspace{-.17cm}
       \includegraphics[width=.22\textwidth]{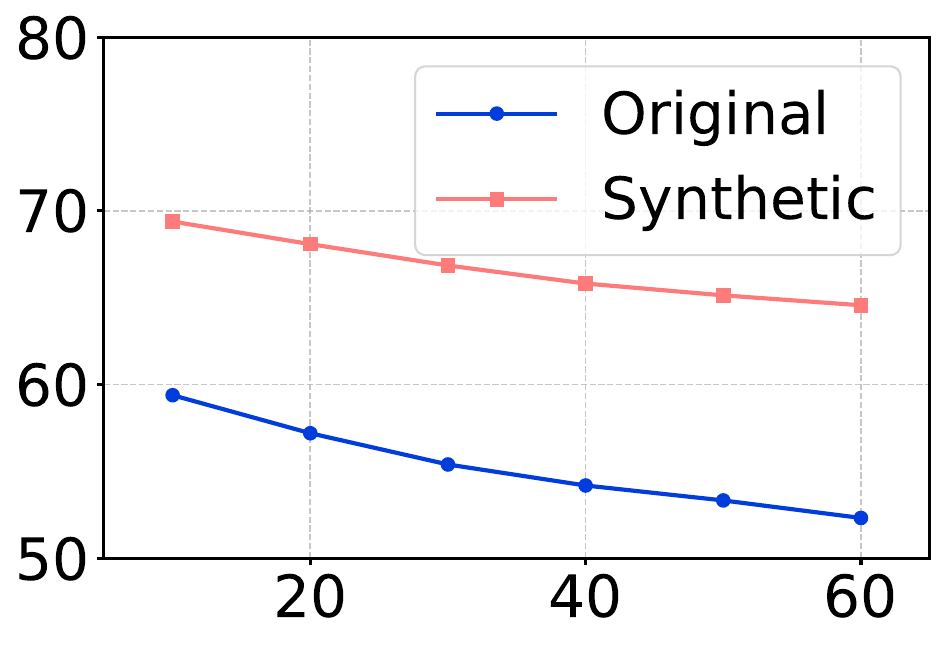} & \hspace{-.17cm}\includegraphics[width=.22\textwidth]{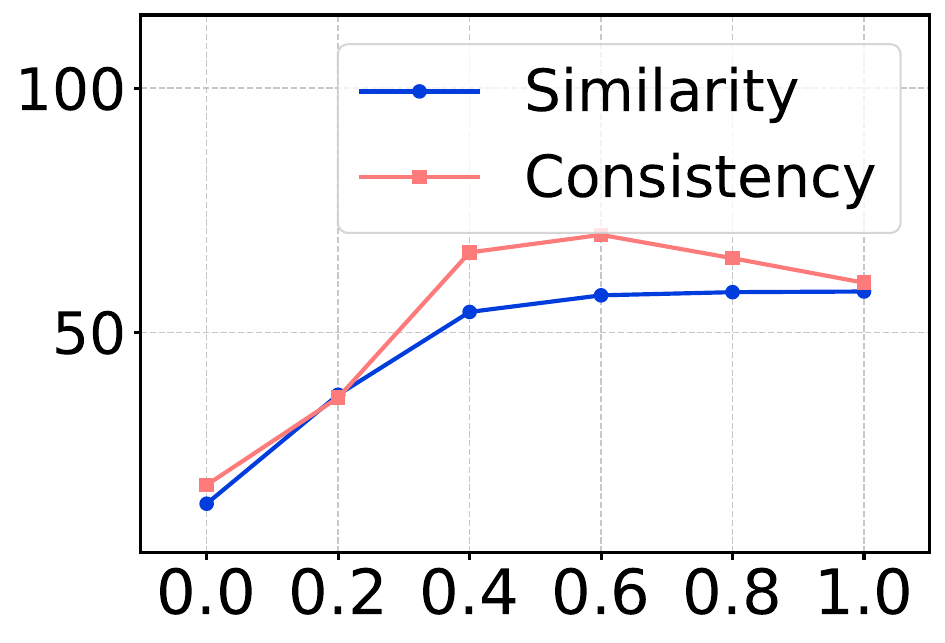}
       \\    
       (a) KNN Sampling & (b) Selection of $\lambda$
       \\
\end{tabular}
  \caption{Experiments (a) and (b) were both conducted on the TREC dataset. Figure (a): KNN sampling with different values of \( K \); Figure (b): Synthetic embeddings using different values of \(\lambda\), with only 8 demonstrations selected.}
  \label{fig:label_consistency}

\end{figure}

\noindent{\bf Role of Parameter \( \lambda \) in Data Synthesis} The efficacy of the TopK-SD method is contingent upon the selection of \(\lambda\). Optimal \(\lambda\) values yield demonstrations that excel in both semantic similarity and label consistency. To elucidate this relationship, we embarked on experiments to assess how varying \(\lambda\) influences these dual metrics as shown in Figure \ref{fig:label_consistency}. In Figure (a), we executed KNN sampling across a spectrum of \(K\) values. Specifically, different values of \(K\), utilizing both the original and synthetic embeddings (\(\lambda = 0.6\)). The synthetic embedding, when subjected to KNN sampling, outperforms the original embedding in procuring demonstrations with congruent labels. As shown in the figure, consistency has essentially improved by more than \(10\%\). Figure (b) examines the average semantic similarity and label consistency of demonstrations from KNN sampling on synthetic embeddings with \(\lambda\) values from 0.0 to 1.0. \(\lambda = 1.0\) is the original embedding. Semantic similarity, measured by cosine similarity (\([-1, 1]\) range, scaled by 100 for clarity), increases with \(\lambda\). Label consistency peaks at \(\lambda = 0.6\). At \(\lambda = 0.6\), semantic similarity is high and stable, while label consistency surges. Our findings underscore the criticality of dataset-specific \(\lambda\) optimization. Further substantiating this point, Table \ref{tab:vote} reveals that the quality of demonstrations selected via different \(\lambda\) values markedly impacts subsequent model inference.

\section{Conclusion}
This work rethinks the learning paradigm of ICL as transductive learning, establishing a transductive label propagation framework based on Bayesian view that highlights the importance of semantic similarity and label consistency to the shared concepts. We propose TopK-SD for demonstration selection. Experiments on multiple benchmarks show significant improvements. It validates the importance of label consistency and confirmed the derivation of Bayesian transduction. We hope this work not only provides a new perspective on the essence of ICL but also offers a robust mathematical foundation for future research.

\section{Acknowledgements}

This work was supported by the National Natural Science Foundation of China (No. U23B2056  and No. 62306026), in part by the National Science and Technology Major Project under Grant 2022ZD0120202, in part by the Fundamental Research Funds for the Central Universities, and in part by the State Key Laboratory of Complex \& Critical Software Environment.
\bibliography{aaai2026}

\end{document}